\documentclass[10pt,twocolumn,letterpaper]{article}

\usepackage{cvpr}              

\usepackage{graphicx}
\usepackage{amsmath}
\usepackage{amssymb}
\usepackage{booktabs}

\usepackage{siunitx}

\usepackage[pagebackref,breaklinks,colorlinks]{hyperref}

\usepackage[capitalize]{cleveref}
\crefname{section}{Sec.}{Secs.}
\Crefname{section}{Section}{Sections}
\Crefname{table}{Table}{Tables}
\crefname{table}{Tab.}{Tabs.}

\def\cvprPaperID{*****} 
\def\confName{CVPR}
\def\confYear{2023}

\begin{document}

\title{Continuous Online Extrinsic Calibration of Fisheye Camera and LiDAR}

\author{Jack Borer
\qquad
Jeremy Tschirner
\qquad
Florian Ölsner
\qquad
Stefan Milz
\\ Spleenlab GmbH
}
\maketitle

\begin{abstract}
Automated driving systems use multi-modal sensor suites to ensure the reliable, redundant and robust perception of the operating domain, for example camera and LiDAR. An accurate extrinsic calibration is required to fuse the camera and LiDAR data into a common spatial reference frame required by high-level perception functions. Over the life of the vehicle the value of the extrinsic calibration can change due physical disturbances, introducing an error into the high-level perception functions. Therefore there is a need for continuous online extrinsic calibration algorithms which can automatically update the value of the camera-LiDAR calibration during the life of the vehicle using only sensor data.  

We propose using mutual information between the camera image's depth estimate, provided by commonly available monocular depth estimation networks, and the LiDAR pointcloud's geometric distance as a optimization metric for extrinsic calibration. Our method requires no calibration target, no ground truth training data and no expensive offline optimization. We demonstrate our algorithm's accuracy, precision, speed and self-diagnosis capability on the KITTI-360 data set.

\end{abstract}

\section{Introduction}
\label{sec:intro}

Sensor fusion of different sensing modalities with disjoint failure distributions, for example camera and LiDAR, is important for high-level safety critical functions found on automated vehicles. Over the life of an automated vehicle, sensor extrinsic calibrations will change due to environmental factors like vibration, temperature and collisions. Even a seemingly innocuous one or two degree error in the value of camera-LiDAR extrinsic calibration can introduce a catastrophic failure into high-level camera-LiDAR perception functions. 

\begin{figure}
    \centering
    \subcaptionbox{Projection with $25^{\circ}$ error}{
        \includegraphics[width =  0.48\linewidth]{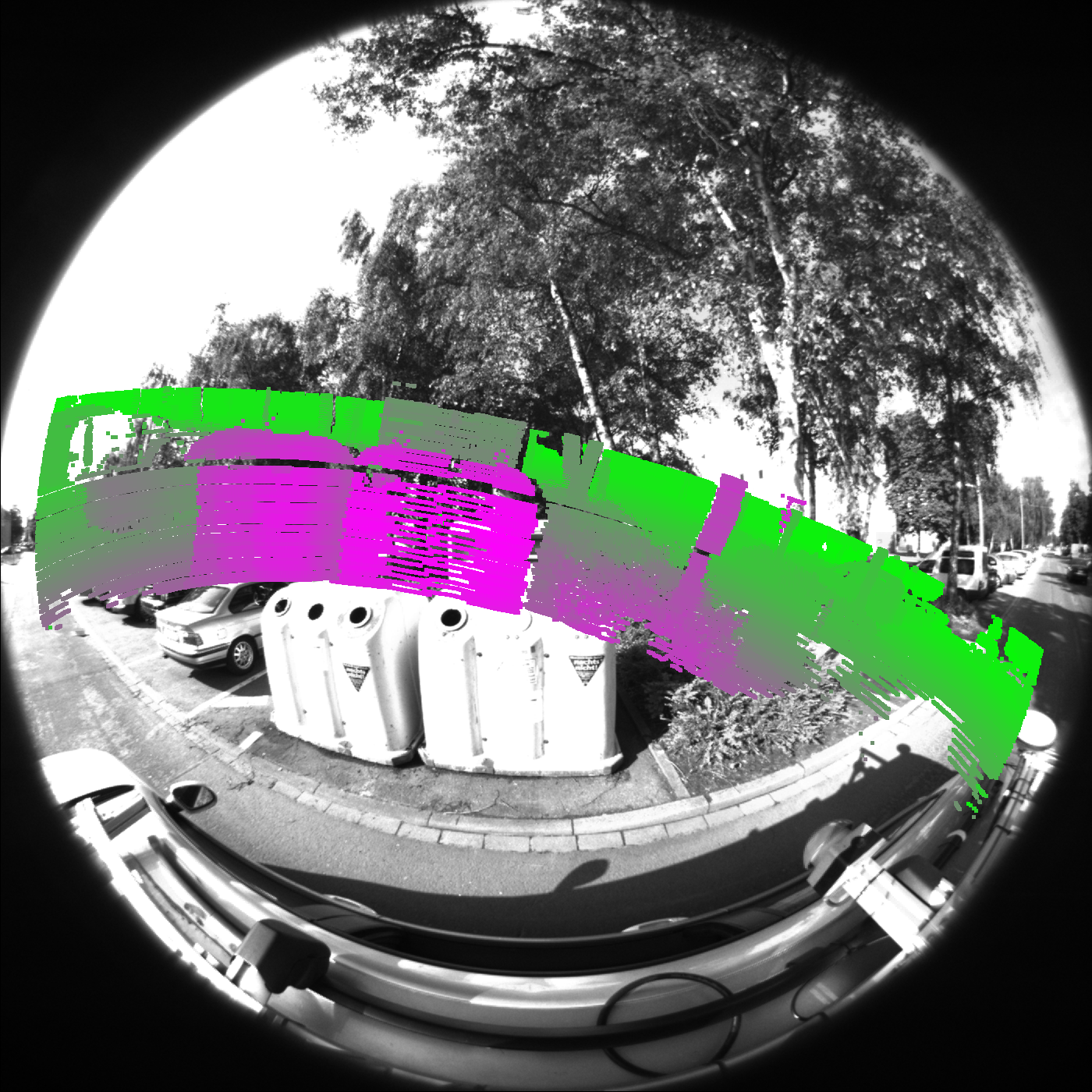}}
            \subcaptionbox{Projection after our COEC}{
        \includegraphics[width =  0.48\linewidth]{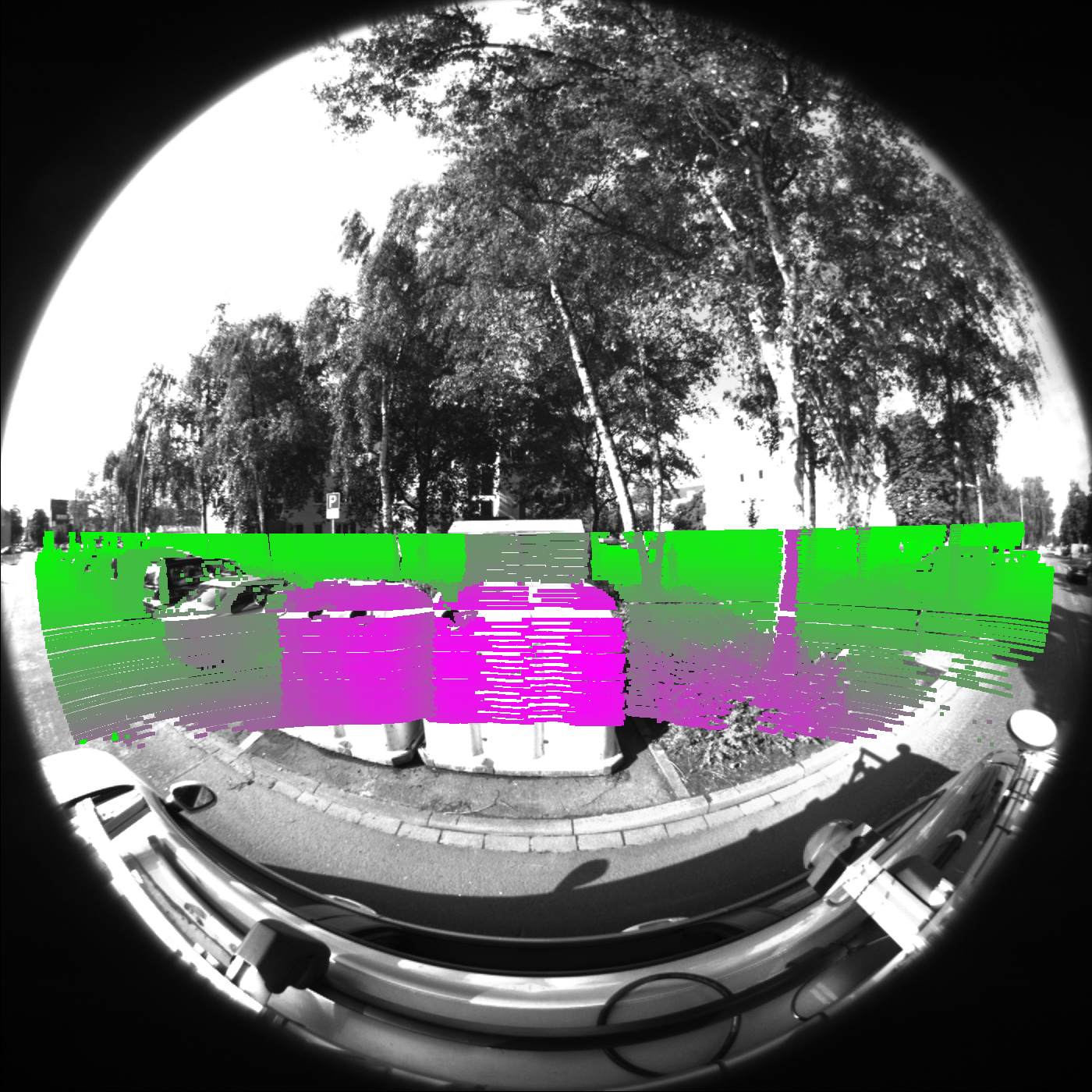}}\quad
    \caption{Incorrect camera-LiDAR extrinsic calibration causes the projection of the pointcloud onto the image to misalign. Shown here is an error input (left) to our algorithm and the optimized result (right).}
    \label{fig:rgb_input_output}
\end{figure}

With advances in automated vehicles, particularly the advent of driver-less vehicles, there is is a growing demand for fail-operational systems. A fail-operational system  can identify a failure, trigger a corrective behavior, and then continue operating unimpaired. This is the context in which continuous online extrinsic calibration (COEC) plays a crucial role in automated vehicle robustness and safety. The ability to re-calibrate the sensor system (see Fig. \ref{fig:rgb_input_output}) after a failure without a in-person vehicle service event is a requirement for fleet scalable safety critical sensor fusion systems. 

We present a lightweight pattern-free online calibration algorithm for the continuous online extrinsic calibration of a fisheye camera and LiDAR using depth feature mutual information (MI). Our main contributions are:

\begin{itemize}
    \itemsep0em 
    \item A end-to-end COEC pipeline using depth feature mutual information
    \item Self-diagnosis confidence metrics for calibration evaluation  
    \item Realistic experiments on the KITTI-360 data set using raw un-rectified fisheye images
\end{itemize}

\section{Related Work}

Feature based methods for camera-LiDAR calibration first extract features from both sensors, then match them and then minimize the reprojection error. Target based approaches \cite{Zhang2004ExtrinsicCO}\cite{Unnikrishnan2005FastEC}\cite{Pandey2010ExtrinsicCO}\cite{Geiger2012AutomaticCA}\cite{Mirzaei20123DLI} require a specifically designed target structure (e.g. a checkerboard) that allows for automatic feature detection. Alternatively, features can be hand-selected and matched by an expert user \cite{Scaramuzza2007ExtrinsicSC}. Both of these methods, although accurate, are unsuitable for COEC.

To address this problem several works \cite{Bileschi2009FullyAC}\cite{Levinson2013AutomaticOC}\cite{Taylor2013AutomaticCO}\cite{Taylor2015MultiModalSC}\cite{MuozBan2020TargetlessCC} propose to correlate edges in camera and LiDAR depth images, because they can be automatically extracted from arbitrary scenes as long as there is sufficient structure. The disadvantage is that edge features on camera data tend to be noisy in highly textured scenes \cite{Zhu2020Online}. Recently, heuristic feature engineering has been replaced by learning based methods \cite{Iyer2018CalibNetGS}\cite{Lv2021LCCNetLA}\cite{Zhao2021CalibDNNMS}\cite{Zhang2022EnhancedLL} that train a neural network to directly predict the calibration parameters given a pair of sensor data. This approach however needs large amounts of expensive well calibrated training data to generalize well.

An entirely different procedure is to treat the LiDAR and camera data as random variables and optimize the mutual information as a measure of their statistical dependence. Pandey \etal \cite{Pandey2012AutomaticTE}\cite{Pandey2015AutomaticEC} applied this first in an automotive setup by assuming a correlation between the projected LiDAR reflectivity and grayscale values from the camera image. Our method expands on this work but uses geometric features instead, provides confidence metrics and is optimized for online application. Furthermore we expand the method to work on un-rectified fisheye camera images using the Double Sphere camera model \cite{Usenko2018Double}.

\section{Method}

\begin{figure}
    \centering
    \subcaptionbox{Projection with $25^{\circ}$ error}{
        \includegraphics[width =  0.48\linewidth]{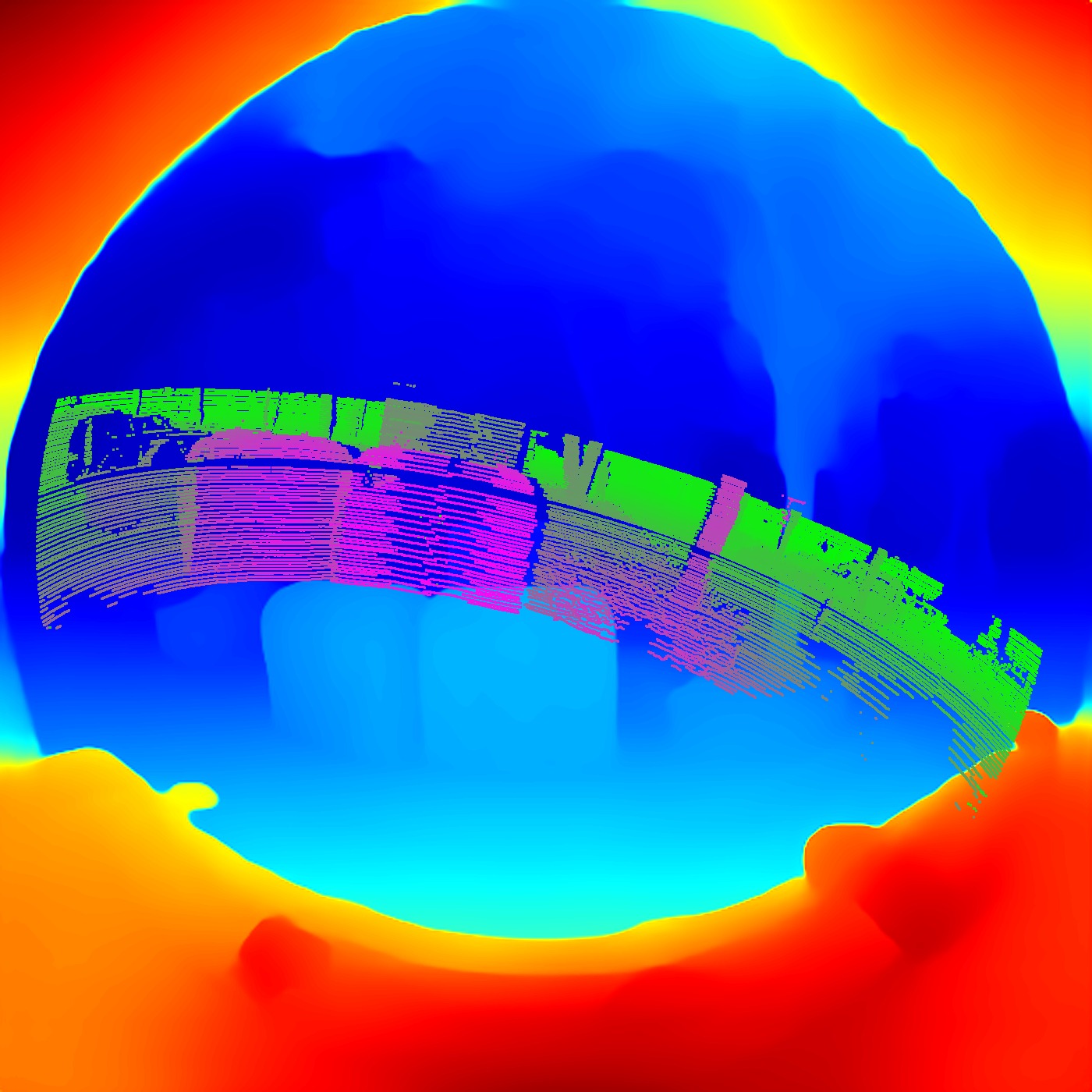}}
            \subcaptionbox{Projection after our COEC}{
        \includegraphics[width =  0.48\linewidth]{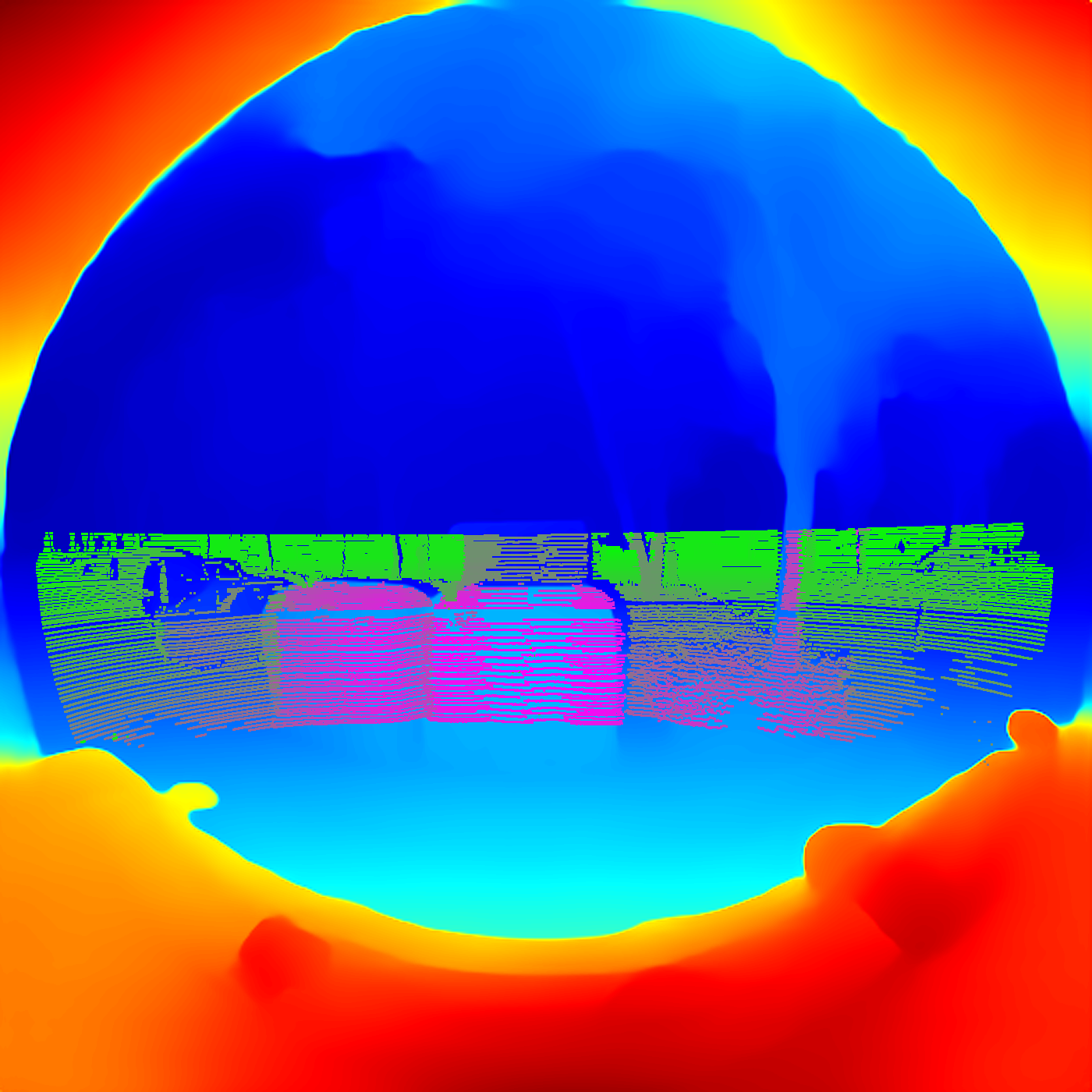}}\quad
    \caption{Our proposed method calculates the mutual information between depth features. Here both the image and pointcloud are colored according to depth to visualize their shared information.}
    \label{fig:monodepth_input_output}
\end{figure}

\begin{figure}
    \centering
    \subcaptionbox{$5^\circ$ Initialization Error}{
        \includegraphics[width =  0.48\linewidth]{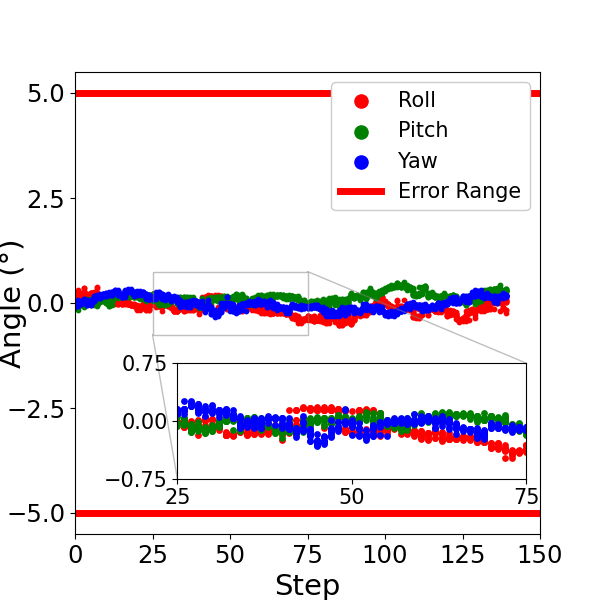}}
            \subcaptionbox{$10^\circ$ Initialization Error}{
        \includegraphics[width =  0.48\linewidth]{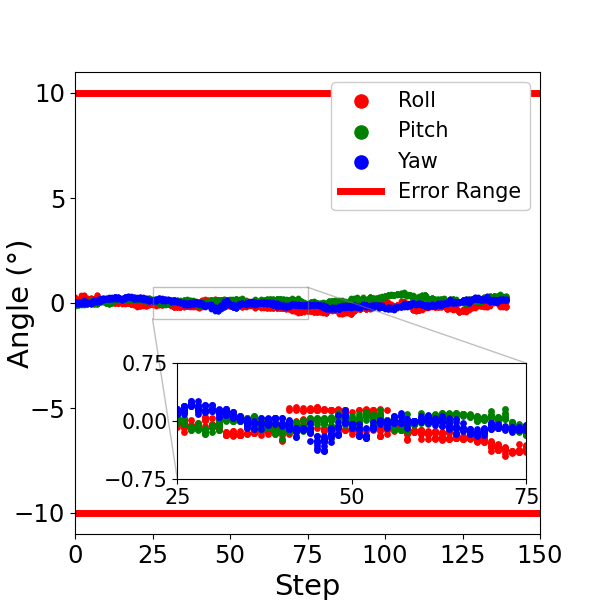}}\quad
    \caption{Extrinsic calibration parameter values calculated every 3 minutes using 25 frames throughout the course of the KITTI-360 driving sequence.}
    \label{fig:coec_rpy_results}
\end{figure}

\begin{figure*}
    \centering
    \subcaptionbox{Mutual Information}{
        \includegraphics[width = 0.325\linewidth]{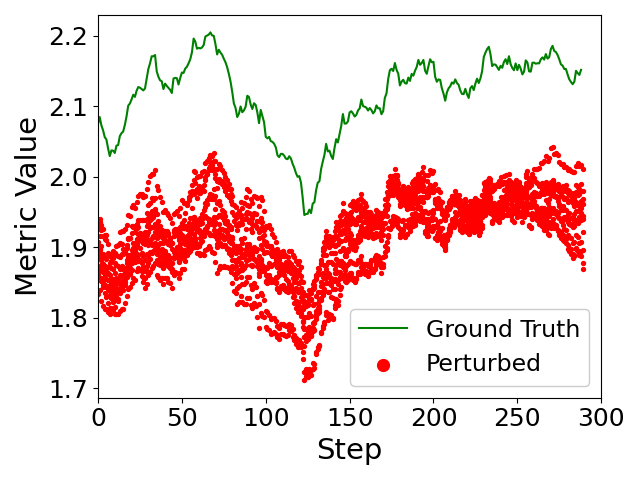}}
            \subcaptionbox{Numerical First Derivative}{
        \includegraphics[width = 0.325\linewidth]{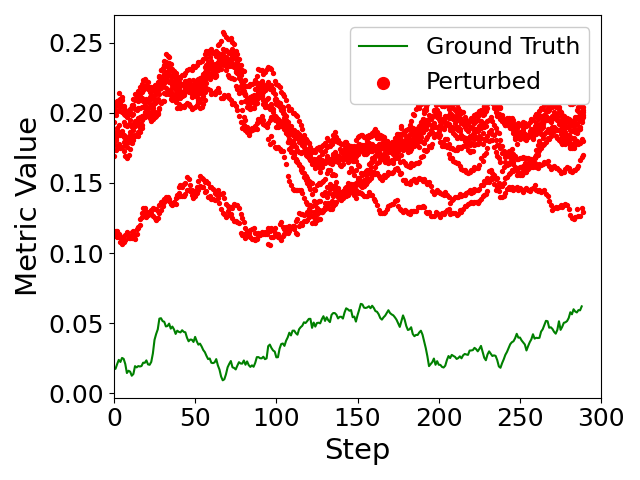}}
    \subcaptionbox{Numerical Second Derivative}{
        \includegraphics[width = 0.325\linewidth]{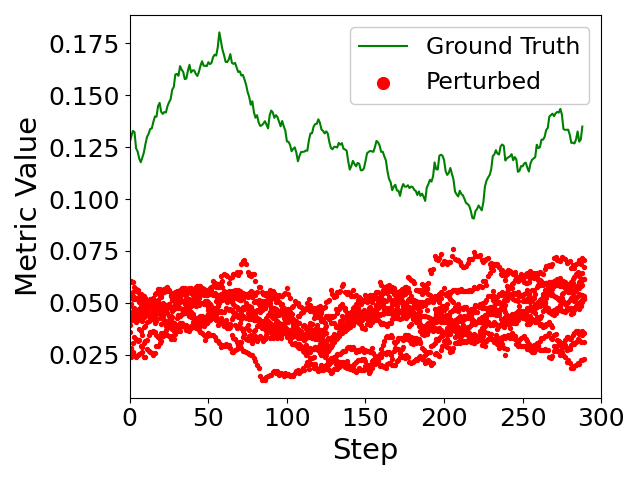}}\quad
    \caption{Failure metric values calculated every 3 minutes using 50 frames throughout the course of the KITTI-360 driving sequence at the ground truth and at a $3^{\circ}$ error. It is apparent that a classifier can be designed to identify failed COEC optimizations using these metrics.}
    \label{fig:failure_metric}
\end{figure*}

Mutual information measures the amount that knowing one variable reduces our uncertainty about the other variable. At the true value of the extrinsic calibration between camera and LiDAR the value of the mutual information should be maximum. This assumes that the image and LiDAR measurement are conditioned on the scene. 

It was previously in \cite{Pandey2012AutomaticTE} demonstrated that image intensity and LiDAR reflectance (an intensity feature) shared information and could be used as a cost metric for camera-LiDAR extrinsic calibration. There is however no theory and little consistency about which scenes share intensity feature information. The brightness (intensity) in an image and the intensity measured by the LiDAR is only loosely correlated. For example, a dark blue and bright white road sign will both return high intensity LiDAR measurements however the image intensity will vary dramatically. Examples such as this, demonstrating the inconsistency between image and LiDAR intensity are common in autonomous driving sensor data.

We propose MI based extrinsic calibration using geometric features. Instead of using the camera intensity directly, we propose considering a third virtual sensor. This virtual depth sensor takes the camera image as input and returns the depth of each pixel in the scene. An extrinsic calibration between the virtual sensor and the LiDAR is also a calibration between the camera and the LiDAR sensor. From the LiDAR pointcloud, instead of using the point intensity we use the euclidean distance of the point from the sensor as the feature (see Fig. \ref{fig:monodepth_input_output}).

The virtual depth sensor is implemented using a monocular depth estimation network. These networks, both self-supervised \cite{Godard2019Monodepth2} and supervised \cite{Ranftl2022MiDaS}, have proliferated in the last five years throughout the research community. Self-supervised networks are exciting because of their ability to extract geometric information from images with no user input. Our proposed calibration algorithm is just one application that takes advantage of this fundamental scientific discovery for the practical purpose of calibrating camera-LiDAR systems.

Understanding if the optimization has converged to the correct value of the extrinsic parameters is equally as important as the optimization itself. We propose three simple metrics taken from the algorithm itself that can be used to identify failure. These are the value of the MI, which is assumed to be maximum at the true value of the extrinsic calibration, the first derivative of the MI with respect to the extrinsic parameters, which should be zero when at a maximum in the information function and the second derivative of the MI with respect to the extrinsic parameters, which reflects the peakedness of the information function and should be maximum at the correct value of the extrinsic calibration. 

\subsection{Algorithm}

The input to our COEC algorithm is a set $\mathcal{F} = \{(I_1, P_1), (I_2, P_2), ..., (I_{N}, P_{N})\}$ of $N$ time synchronized image pointcloud pairs. Each pair consists of an RGB image $I_i$ recorded in the 2D camera optical coordinate frame $\boldsymbol{O}_{CO}$ and a pointcloud $P_i$ including $K$ 3D points recorded in the LiDAR coordinate frame $\boldsymbol{O}_{L}$.  

In the first step we extract a depth map $D_i$ from each RGB image using a pretrained monocular depth estimation network. In the second step we use the rigid transformation ${^C}{}{\boldsymbol{T}_L} \in \mathbb{R}^{4\times4}$ parameterized by an initial guess $\boldsymbol{\Theta}_0 = [\theta_{x}, \theta_{y}, \theta_{z}, t_{x}, t_{y}, t_{z}]^T$ to first transform each pointcloud into the 3D camera coordinate frame $\boldsymbol{O}_{C}$ and further project it into the camera optical coordinate frame $\boldsymbol{O}_{CO}$. The projection is done based on a double sphere camera model and the projective camera matrix  $\boldsymbol{P} \in \mathbb{R}^{3\times4}$.

As a result we get for each depth map pointcloud pair $\{D_i, P_i\}$ a set of $M \leq K$ pixels onto which the 3D LiDAR points are projected. From the pixels we can directly extract the correspondent image depth features $f^D = \{d_1^D, d_2^D, ..., d_M^D\}$ by taking the respective values from the depth map. The matching pointcloud depth features $f^P = \{d_1^P, d_2^P, ..., d_M^P\}$ are calculated using the euclidean distance to the LiDAR coordinate frame origin from each of the $K$ 3D points. With both feature sets we approximate their MI using a normalized histogram representation:
\begin{equation}
  MI(f^P, f^D, \boldsymbol{\Theta}) = H(f_{\boldsymbol{\Theta}}^P) + H(f_{\boldsymbol{\Theta}}^D)  - H(f_{\boldsymbol{\Theta}}^P, f_{\boldsymbol{\Theta}}^D)
\end{equation}

The MI value directly depends on the extrinsic parameters $\boldsymbol{\Theta}$. For a more robust estimate we average the MI value over the whole input set. Finally, we use the average MI as an objective function parameterized by the extrinsic parameters and maximize it to yield the correct extrinsic calibration parameters. The optimization is done with Powell's BOBYQA algorithm \cite{bobyqa} for solving bounded optimization problems without derivatives.

\section{Experiments}

The following experiments were performed on the 28/05/2013 KITTI-360 drive 0000 synchronized data. It contains 11,518 synchronized sensor frames at \qty{10}{\hertz} which corresponds to about 20 minutes and is predominantly in a suburban environment. Pointclouds were motion compensated using the provided ego-motion and timing data. Monodepth predictions were performed using the MiDaS \cite{Ranftl2022MiDaS} dpt\_beit\_large\_512 network. This network was chosen in the interest of reproducibility. Although MiDaS is not strictly self-supervised monodepth, it is a publicly available set of networks that perform adequately on fisheye images.

\subsection{Continuous Online Extrinsic Calibration}
In this experiment we test the ability of our proposed COEC algorithm to calculate the correct value of the rotation component of the extrinsics. At each step we use 25 image-pointcloud pairs taken over a three minute driving slice. Then we run COEC ten times with errors initialized from a sphere with radius $5^{\circ}$ and $10^{\circ}$. We move forward through the sequence by an increment of 72 frames after each step. The results of the experiment are shown in Tab. \ref{table:coec_rpy_results} and Fig. \ref{fig:coec_rpy_results}. The average error from the adjusted ground truth is under $0.2^{\circ}$ and there is not a single catastrophic failure. 

\begin{table}
\small
\addtolength{\tabcolsep}{-1pt}
\begin{center}
    \begin{tabular}{|c|c|c|}
    \hline
    Error & $5^\circ$ & $10^\circ$ \\
    \hline 
    Roll & $-0.118 \pm 0.175$ & $-0.121 \pm 0.171$ \\
    Pitch & $0.129 \pm 0.117$ &  $0.130 \pm 0.117$ \\
    Yaw & $-0.010 \pm 0.151$  & $-0.011 \pm 0.151$  \\
    \hline
    \end{tabular}
\end{center}
\caption{Mean extrinsic parameter rotation values in degrees for 1400 COEC executions at an initial error of $5^{^\circ}$ and $10^{^\circ}$. Our methods robustness to initialization error is apparent by the fact that for both initialization ranges the optimized value is nearly identical.}
\label{table:coec_rpy_results}
\end{table} 

\begin{table}
\small
\addtolength{\tabcolsep}{-1pt}
\begin{center}
\begin{tabular}{|c|c|c|}
\hline
Frames & \begin{tabular}[c]{@{}c@{}}Calculate\\ MI (ms)\end{tabular} & \begin{tabular}[c]{@{}c@{}} Optimization (ms)\end{tabular} \\ \hline
5 & $4.2 \pm 0.9$ & $808.1 \pm 297.3$ \\
25 &  $21.5 \pm 1.4$ & $4221.8 \pm 1356.3$ \\
50 & $42.8 \pm 1.6$ & $7915.4 \pm 2039.6$ \\ \hline
\end{tabular}
\end{center}
\caption{Runtime results for calculating the MI and for the complete COEC optimization on different sized frame sets. The required time scales linearly with the number of frames used.}
\label{table:coec_runtime_results}
\end{table}

\subsection{Confidence Metrics}
In this experiment we demonstrate that our proposed COEC algorithm produces metrics which can be used to classify success and failure. We present results for the values of three metrics; mutual information, numerical first derivative and numerical second derivative. We use 50 image-pointcloud pairs taken over a three minute driving slice to calculate each metric. The value of the metrics at the ground truth value of the calibration and the value of the metric at an error value of $1^{\circ}$ and $3^{\circ}$ is shown in Fig. \ref{fig:failure_metric}. We define a simple classifier $C$ that using a threshold value of the three metrics ($MI_{lim}, f'_{lim}, f''_{lim}$) can classify a extrinsic parameter set as a failure,

\begin{equation}\label{model3_coef}
    \resizebox{0.9\hsize}{!}{%
        $ C(\Theta) = \begin{cases} \mbox{true,} & MI(\Theta) < MI_{lim} \mbox{ or }  f'(\Theta) > f'_{lim} \mbox{ or }  f''(\Theta) < f''_{lim} \\ \mbox{false,} & \mbox{otherwise} \end{cases} $%
        }
\end{equation}

With this simple metric are able to classify 99.8\% and 100\% of parameter sets $\Theta$ at $1^{\circ}$ and $3^{\circ}$ error as failures.

\subsection{Identifying Failures}
\begin{figure}
    \centering
    \includegraphics[width = 0.75\linewidth]{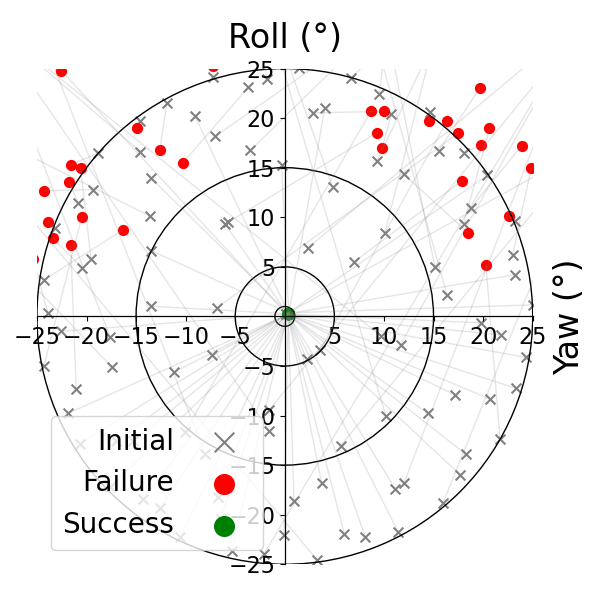}
    \caption{Failed optimizations identified using our proposed metric threshold plotted in a bullseye plot. All 36 success and 64 failures were properly classified using our proposed metrics.}
    \label{fig:failure_experiment}
\end{figure}

In this experiment we test the ability of our proposed COEC algorithm metrics to self-diagnose the failure of an example optimization. We use the metric threshold values to classify the results of 100 executions of our COEC algorithm on a single 25 frame three minute sequence. The results of this experiment are shown in Fig. \ref{fig:failure_experiment}. Out of 100 executions initialized with a $25^{\circ}$ 36 executions converge to an error of less than $0.5^{\circ}$ and 64 fail to converge. Our threshold metrics allow the correct classification of all failures and success.

\subsection{Runtime Performance}
The runtime performance experiments were conducted on an NVIDIA Jetson Orin with a 12-core 1.3 GHz CPU and an Ampere GPU with 2048 CUDA cores. Each result is obtained on the same KITTI-360 sequence as before and averaged over 100 trials. Table \ref{table:coec_runtime_results} shows mean and standard deviation of the measured time
for just calculating the average MI and the full optimization. The results show that the computation time grows linearly with the amount of frames taken and that the algorithm is suitable to run live on an embedded system. Although the depth maps were pre-computed in our case, the MiDaS authors state that the dpt\_beit\_large\_512 network runs at 5.7 FPS on a RTX3090 GPU.

\section{Conclusion}

We proposed a continuous online extrinsic calibration algorithm using the mutual information between a camera monocular depth estimate and LiDAR depth feature. We demonstrate our method's accuracy, precision, self-diagnosis ability and timing performance which make it suitable for automotive COEC applications.

Future work can include,
\begin{itemize}
    \item Exploration of the statistical properties of monocular depth estimation and LiDAR mutual information
    \item Understanding the relationship to the number of data used
    \item More comprehensive confidence metrics and robust classifiers 
    \item Expansion to more camera and LiDAR models
\end{itemize}

\clearpage
{\small
\bibliographystyle{ieee_fullname}
\bibliography{egbib}
}

\end{document}